\documentclass[11pt,a4paper]{article}
\usepackage[hyperref]{acl2021}
\usepackage{times}
\usepackage{latexsym}

\usepackage{microtype}
\usepackage{booktabs}
\usepackage[hyperref]{acl2021}
\usepackage{times}
\usepackage{latexsym}
\usepackage{amsmath}
\usepackage{amsthm}
\usepackage{eucal}
\usepackage{amssymb}
\usepackage{mathrsfs}

\usepackage{multirow}
\usepackage{graphics}
\usepackage{amssymb}
\usepackage{pifont}
\newcommand{\cmark}{\ding{51}}%
\newcommand{\xmark}{\ding{55}}%
\usepackage{microtype}
\usepackage[colorinlistoftodos]{todonotes}
\usepackage[belowskip=0pt,aboveskip=5pt,font=small]{caption}
\aclfinalcopy 

\title{Zero-shot Medical Entity Retrieval without Annotation: Learning From Rich Knowledge Graph Semantics}

\author{Luyang Kong \\
  Amazon AI \\
  \texttt{luyankon@amazon.com} \\
  \And
  Christopher Winestock \\
  Amazon AI \\
  \texttt{winestoc@amazon.com} \\
  \And
  Parminder Bhatia \\
  Amazon AI \\
  \texttt{parmib@amazon.com} \\}

\date{}

\begin{document}
\maketitle

\begin{abstract}
Medical entity retrieval is an integral component for understanding and communicating information across various health systems. 
 Current approaches tend to work well on specific medical domains but generalize poorly to unseen sub-specialties. 
 This is of increasing concern under a public health crisis as new medical conditions and drug treatments come to light frequently.
 Zero-shot retrieval is challenging due to the high degree of ambiguity and variability in medical corpora, making it difficult to build an accurate similarity measure between mentions and concepts. 
 Medical knowledge graphs (KG), however, contain rich semantics including large numbers of synonyms as well as its curated graphical structures. 
 To take advantage of this valuable information, we propose a suite of learning tasks designed for training efficient zero-shot  entity retrieval models. 
 Without requiring any human annotation, our knowledge graph enriched architecture significantly outperforms common zero-shot benchmarks including BM25 and Clinical BERT with $7\%$ to $30\%$ higher recall across multiple major medical ontologies, such as UMLS, SNOMED and ICD-10. 

\end{abstract}


\section{Introduction}
Entity retrieval is the task of linking mentions of named entities to concepts in a curated knowledge graph (KG). 
This allows medical researchers and clinicians to search medical literature easily using standardized codes and terms to improve patient care.
Training an effective entity retrieval system often requires high quality annotations, which are expensive and slow to produce in the medical domain. 
It is therefore not feasible to annotate enough data to cover the millions of concepts in a medical KG, and difficult to adapt quickly enough to those newly appeared medical conditions and drug treatments under a public health crisis.
Hence, a robust medical entity retrieval system is expected to have decent performance in a zero-shot scenario. 

Zero-shot retrieval is challenging due to the complexity of medical corpora - large numbers of ambiguous terms, copious acronyms and synonymous terms.
It is difficult to build an accurate similarity measure which can detect the true relatedness between a mention and a concept even when their surface forms differ greatly. 

Early entity retrieval systems use string matching methods such as exact match, approximate match \cite{prominer} and weighted keyword match e.g.
BM25 \cite{WANG2020103418}. 
Although annotated training data is not required, such systems typically lack the ability to handle synonyms and paraphrases with large surface form differences. 
In recent years, large scale pretraining \cite{bert} has been widely adopted in the medical domain such as Clinical BERT \cite{clinical_bert} and BioBERT \cite{biobert}. 
\citet{Snomed2Vec-DSHealth2019} also integrates graph structure information during pretraining. 
Most of them, however, require a finetuning step on annotated training data \cite{wu2019zero} before being applied to entity retrieval. 

As an alternative to manually annotating a corpus, the rich semantics inside a KG itself can be utilized \cite{chang2020pretraining}. 
One important entry is the \textbf{synonym}, whereby two medical terms may be used interchangeably. 
In addition, the \textbf{graphical structure} of a KG  contains information on how concepts are related to each other and so can be used as another valuable resource for building an effective similarity measure. We therefore design \emph{synonym-based tasks} and \emph{graph-based tasks} to mine a medical KG. Trained with our proposed tasks, a simple Siamese architecture significantly outperforms common zero-shot benchmarks across multiple major medical ontologies including UMLS, SNOMED and ICD10.

Our contributions are as follows. 
(1) We propose a framework which allows the information in medical KGs to be incorporated into entity retrieval models, thereby enabling robust zero-shot performance without the need of human annotations. 
(2) We apply the framework to major medical ontologies and conduct extensive experiments to establish the effectiveness of our framework.
(3) When annotations are available, we show that the proposed framework can be easily plugged into an existing supervised approach and in so doing, deliver consistent improvements.

\section{Formulation}
\textbf{Entity retrieval}. Given a \emph{mention} 
$\boldsymbol{m}$
and a \emph{concept} 
$\boldsymbol{c} \in KG = \{c_1, c_2, ..., c_n\}$, the goal is to learn a similarity measurement
$S(\boldsymbol{m},\boldsymbol{c})$
, so that the most relevant concept is assigned the highest score. 
A concept is also referred to as a \emph{node} in a KG. 
We use them interchangeably below.

\textbf{Zero-shot entity retrieval}. 
We examine two zero-shot scenarios: 1) \emph{zero-shot on mentions only}, which assumes unseen mentions but allows seen concepts at test time. 2) \emph{zero-shot on mentions and concepts}, which assumes both to be unseen at test time.

\section{Model Architecture}

\textbf{Siamese architecture}. 
Mention $\boldsymbol{m}$
and concept $\boldsymbol{c}$
are firstly embedded into vectors, using a shared function $T$: $\vec{e_m} = T(\boldsymbol{m}), \vec{e_c} = T(\boldsymbol{c})$.
$T$ is also referred to as an encoder, for which we use the Transformer \cite{transformer} encoder in this work. 
Similarity between a mention and a concept is then measured as the inner product:
$
S(\boldsymbol{m}, \boldsymbol{c}) = 
\langle\, \vec{e_m}, \vec{e_c} \rangle
$.
\medskip

\textbf{Optimization}. Assume model parameter is $\theta$. We use in-batch negatives for optimization. Loss function for a batch of size $B$ is defined as mean negative log likelihood:
\[L = -\frac{1}{B}
    \sum_{i=1}^{B} 
    \log (
        P((\boldsymbol{m_i},\boldsymbol{c_i})|\theta) 
        )
\]
where the conditional probability of each mention-concept pair $(\boldsymbol{m_i},  \boldsymbol{c_i})$ in the batch is modeled as a softmax:
\[
P((\boldsymbol{m_i},\boldsymbol{c_i})|\theta) = 
    \frac{\exp(S_\theta(\boldsymbol{m_i}, \boldsymbol{c_i}))}
         {\sum_{j=1}^{B}\exp(S_\theta(\boldsymbol{m_j}, \boldsymbol{c_j}))}
\]

\section{Learning Task}
We design our learning tasks by constructing mention-concept pairs 
$(\boldsymbol{m},  \boldsymbol{c})$. 
The goal is to capture multiple layers of semantics from a KG by leveraging its unique structure. 
Since each structure implies its own measure of similarity, we design learning tasks by finding very similar or closely related textual descriptions and use them to construct
$(\boldsymbol{m},  \boldsymbol{c})$
pairs. 
We define two major types of tasks: \emph{synonym-based} tasks and \emph{graph-based} tasks. These are illustrated below for three major medical KG: ICD-10, SNOMED and UMLS.

\subsection{ICD-10}
The 10th version of the International Statistical Classification of Diseases, Clinical Modification (ICD-10) is one of the most widely used terminology systems for medical conditions. It contains over 69K concepts, organized in a tree structure of parent-child relationships.

\begin{figure}[t]
\vskip 0.2in
\begin{center}
\centerline{\includegraphics[width=1.025 \columnwidth]{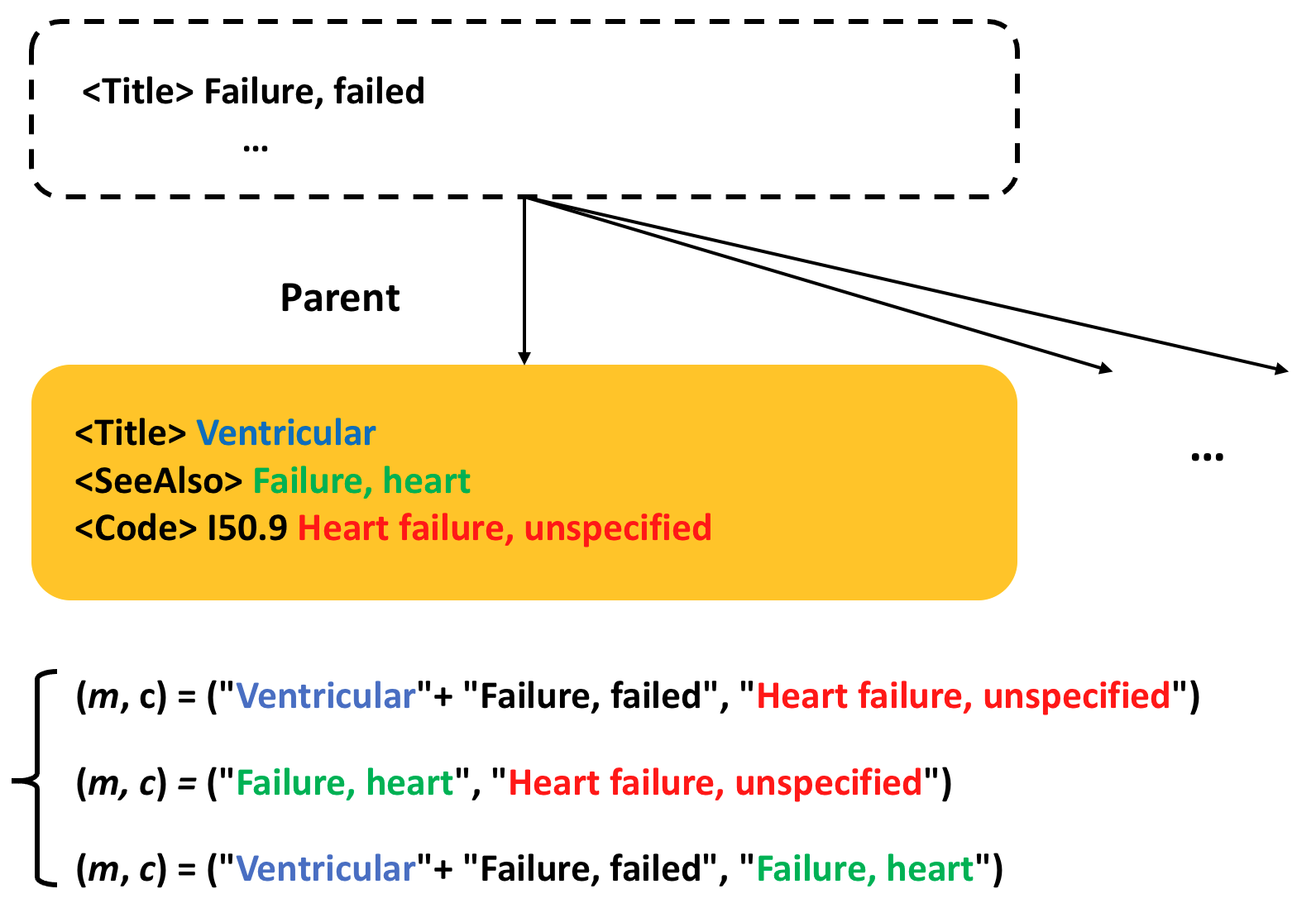}}
\caption{ICD-10 synonym-based task defined at an example node}
\label{icd-syn-tsk}
\end{center}
\vskip -0.15in
\end{figure}

\textbf{Synonym-based task}. In ICD-10, a child node is a more specific medical condition compared to its parent (e.g. \texttt{R07.9 Chest pain, unspecified} is a child of \texttt{R52 Pain, unspecified}). 
Each node $N_i$ has three sections:
The \emph{Title} section contains a subspecifier  (e.g. \texttt{Chest}) of the title of the parent (e.g. \texttt{Pain}), therefore their concatenation gives the full concept description (e.g. \texttt{Chest Pain}). We denote it by $N_{i}^{TitleConcatenation}$. 
The \emph{Code} section contains an ICD-10 code and its formal medical definition, denoted by $N_{i}^{CodeDescription}$.
The \emph{SeeAlso} section contains a similar concept, denoted by $N_{i}^{SeeAlso}$.

These three sections describe the same medical condition with different surface forms, therefore we define the ICD-10 synonym-based task as:
\[\boldsymbol{m} = N_i^{L}, \boldsymbol{c} = N_i^{R}\]
\[ N_i^{L}, N_i^{R} \in \{N_{i}^{TitleConcatenation}, \]
\[N_{i}^{CodeDescription}, N_{i}^{SeeAlso}\}, L \neq R \]

We illustrate it with an example in Figure \ref{icd-syn-tsk}. 

\textbf{Graph-based task}. To incorporate the semantics of parent-child relationships into learning, we define ICD-10 graph-based task as:
\[
\boldsymbol{m} = N_i^{CodeDescription}, 
\boldsymbol{c} = N_j^{CodeDescription}
\]
\[
N_i.is\_parent(N_j)
\]

\subsection{SNOMED}
Systematized Nomenclature of Medicine -- Clinical Terms (SNOMED) is a standardized clinical terminology used for the electronic exchange of clinical health information with over 360K active concepts. 

\textbf{Synonym-based task}. Each node $N_i$ in SNOMED has multiple synonymous descriptions $\{l_i^{1}, l_{i}^{2}, ..., l_{i}^{d}\}$, with $l_i^{1}$ as the main description.
We therefore define SNOMED synonym-based task as:
\[\boldsymbol{m}=l_i^{p}, 
\boldsymbol{c} = l_i^{q}, 
p>q\] 

$\frac{d*(d-1)}{2}$ unique 
$(\boldsymbol{m},  \boldsymbol{c})$
pairs are constructed at each node.

\textbf{Graph-based task}. SNOMED is a directed graph with 107 possible relationship types (e.g. \texttt{is\_a}, \texttt{finding\_site}, \texttt{relative\_to}). A direct connection between two nodes is likely to imply a certain degree of similarity, thus we define the SNOMED graph-based task as:
\[\boldsymbol{m}=l_i^{1}, 
\boldsymbol{c} = l_j^{1}\] 
\[ N_i.is\_connected(N_j) \]

\subsection{UMLS}
The Unified Medical Language System (UMLS) is a compendium of a large number of curated biomedical vocabularies with over 1MM concepts. 
UMLS has almost the same structure as SNOMED, therefore we define the \textbf{synonym-based task} and \textbf{graph-based task} in a similar fashion to that of SNOMED. 

\begin{table}[t]
    \centering
    \small \addtolength{\tabcolsep}{-0pt}
    \begin{tabular}{l l|r|r}
        KG &Task Type& Train & Dev \\
        \toprule
        \multirow{2}{*}{ICD-10} 
        & syn & 113K & 28K \\
        & graph & 33K & 8K \\
        \hline
        \multirow{2}{*}{SNOMED} 
        & syn & 1.4M & 374K \\
        & graph & 955K & 238K \\
        \hline
        \multirow{2}{*}{UMLS} 
        & syn & 27M & 7M \\
        & graph & 7M & 2M \\
        \hline
        \multicolumn{2}{c|}{Comb} & \multirow{2}{*}{198K} & \multirow{2}{*}{488K} \\
        \multicolumn{2}{c|}{(\emph{by down-sampling})} && \\
        \toprule
    \end{tabular}
    \caption{\textbf{Task Description}: Number of $(\boldsymbol{m},\boldsymbol{c})$ pairs in train and dev for all tasks.}
    \label{tab:tasks}
\end{table}

\begin{table}[t]
    \centering
    \small \addtolength{\tabcolsep}{-0pt}
    \begin{tabular}{l l|r|r}
        Dataset &Split& KG & Test size\\
        \toprule
        MedM.& - & UMLS & 66,572 \\
        \hline
        \multirow{4}{*}{COMETA} 
        & SG & \multirow{4}{*}{SNOMED} & 4,350 \\
        & SS & & 4,369 \\
        & ZG & & 3,995 \\
        & ZS & & 4,283 \\
        
        \hline
        \multirow{3}{*}{3DNotes} &ICD& ICD-10 & 5,742 \\
        &SN& SNOMED & 7,521\\
        \toprule
    \end{tabular}
    \caption{\textbf{Test Set Size.}}
    \label{tab:dataset}
\end{table}

\tabcolsep=0.15cm
\begin{table*}[t]
    \footnotesize
    \centering
    \begin{tabular}{l l l || r | r | r| r| r| r| r| r| r}
        \multirow{3}{*}{ Dataset } &
        \multirow{3}{*}{ Split } &
        \multirow{3}{*}{ KG } &
        \multirow{3}{*}{ BM25 } &
        \multirow{3}{*}{ Clinical } &
        \multicolumn{7}{c}{Siamese architecture trained with KG learning tasks (\textbf{ours})} \\
        &&&&& \multicolumn{2}{c|}{ICD-10}&
        \multicolumn{2}{c|}{SNOMED}&
        \multicolumn{2}{c|}{UMLS} &
        \multirow{2}{*}{ Comb } \\
        &&&& BERT & \multicolumn{1}{r|}{Syn} & \multicolumn{1}{r|}{Graph}& 
             \multicolumn{1}{r|}{Syn} & \multicolumn{1}{r|}{Graph}& 
             \multicolumn{1}{r|}{Syn} & \multicolumn{1}{r|}{Graph} &\\
        \toprule
        MedM. & - &UMLS&.04(.17)&.10(.30)&.31(.58)&.31(.56)&.32(.55)&.33(.61)&\textbf{.32(.53)}&\textbf{.30(.57)} & \textbf{.32(.60)}\\
        \hline
        \multirow{4}{*}{COMETA} 
        & SG & \multirow{4}{*}{SNOMED} &.02(.10)&.01(.06)&.30(.52)&.30(.48)&\textbf{.43(.65)}&\textbf{.37(.58)}&.33(.50)&.32(.54)&\textbf{.37(.58)} \\
        & SS &&.02(.11)&.01(.06)&.28(.51)&.28(.47)&\textbf{.41(.62)}&\textbf{.36(.56)}&.31(.48)&.31(.52)& \textbf{.35(.56)}\\
            & ZG &&.02(.12)&.01(.07) &.32(.57)&.32(.54)&\textbf{.47(.71)}&\textbf{.39(.61)}&.36(.55)&.33(.57)&\textbf{.40(.62)} \\
        & ZS &&.02(.10)&.01(.07)&.30(.52)&.29(.47)&\textbf{.40(.64)}&\textbf{.35(.57)}&.31(.49)&.29(.53)&\textbf{.35(.57)} \\
        \hline
        \multirow{2}{*}{3DNotes} & ICD &ICD-10&.05(.22)&.11(.17)&\textbf{.28(.54)}&\textbf{.23(.46)}&.20(.45)&.20(.52)&.18(.39)&.21(.53)&\textbf{.30(.54)} \\
        & SN &SNOMED&.07(.20)&.01(.05)&.20(.50)&.18(.45)&\textbf{.38(.63)}&\textbf{.25(.61)}& .25(.49)& .29(.55)&\textbf{.34(.59)}\\
        \toprule
    \end{tabular}
    \caption{Retrieval performance R@1(25). Siamese architecture trained with our tasks are shown to significantly outperform benchmarks. Evaluation for \emph{zero-shot on mentions only} is  highlighted in \textbf{bold} the rest belongs to \emph{zero-shot on mentions and concepts}. The former enjoys a bigger gain as expected.  }
    \label{tab:results}
\end{table*}

\begin{table}[t]
    \centering
    \small \addtolength{\tabcolsep}{-0pt}
    \begin{tabular}{l l c c}
        Mention & Gold Concept & Syn & Graph \\
        \toprule
        shortness of breath & dyspnea (finding) & \cmark & \xmark \\
        \hline
        \multirow{2}{*}{GI hemorrhage}  & gastrointestinal & 
        \multirow{2}{*}{\cmark} & \multirow{2}{*}{\xmark} \\
        & hemorrhage (disorder) & & \\
        \hline
        \multirow{2}{*}{coronary structure}& 
        coronary artery& 
        \multirow{2}{*}{\xmark} & \multirow{2}{*}{\cmark} \\
        & (body structure) &&\\
        \hline
        \multirow{2}{*}{heart} 
        & heart structure  & 
        \multirow{2}{*}{\xmark} & \multirow{2}{*}{\cmark} \\
        &(body structure) &&\\
        \toprule
    \end{tabular}
    \caption{Prediction error of the model trained with SNOMED tasks evaluated on 3DNotes-SN.}
    \label{tab:case_review}
\end{table}

For each task mentioned above, 
the $(\boldsymbol{m},  \boldsymbol{c})$
pairs generated at each node are combined and split into train and dev in a $80$:$20$ ratio. We also define a \textbf{comb task}, where all the tasks are firstly down-sampled to equal sizes and then combined. A summary can be found in Table \ref{tab:tasks}. 

\section{Datasets}
We include three datasets in zero-shot evaluations. 
{\bf MedMention} \cite{medmention_dataset} is a publicly available corpus of 4,392 PubMed\footnote{https://www.ncbi.nlm.nih.gov/pmc/} abstracts with biomedical entities annotated with UMLS concepts. 
{\bf COMETA} \cite{cometa} is one of the largest public corpora of social media data with SNOMED annotations. 
It provides four train, dev, test splits: Stratified-General (SG), Stratified-Specific (SS), Zeroshot-General (ZG), Zeroshot-Specific (ZS). 
We also use a de-identified corpus of dictated doctor’s notes named {\bf 3DNotes}\cite{LATTE}. 
It has two sets of annotations: one with ICD-10 (ICD split), another with SNOMED (SN split). The annotation follows the i2b2 challenge \cite{i2b2} guidelines. 

Zero-shot performance is evaluated on the corresponding test sets. We report sizes of the test sets in Table \ref{tab:dataset}.

\section{Experiments}
\subsection{Experimental Settings}

We train a model using each task and evaluate them across all test sets. Performance is compared against two common zero-shot entity retrieval benchmarks: BM25, and BERT encoder followed by inner-product similarity. 
For the latter, we tested multiple pre-trained versions including BERT\textsubscript{base}, Clinical BERT and BioBERT. 

\textbf{Hyperparameters}. For our Siamese architecture, the transformer encoder is initialized with BERT\textsubscript{base}. 
We use the BertAdam optimizer with a batch size of $128$, the initial learning rate of $3\times10^{-5}$, warm-up ratio of $0.02$, max epochs of $50$, followed by a linear learning rate decay. 

\textbf{Evaluation metrics}. Top-$k$ retrieval recalls (R@1, R@25) are used as metrics. We also assume that each mention has a valid gold concept in the KG.

\subsection{Results}
We report overall results in Table \ref{tab:results}. 
Clinical BERT consistently outperforms the other pre-trained counterparts, which are therefore omitted. For evaluations of
\emph{zero-shot on mentions only} (e.g. UMLS tasks evaluated on MedMention which is UMLS annotated), we observe $12\%$ to $45\%$ gain for R@1 compared to benchmarks. 
For evaluations of \emph{zero-shot on mentions and concepts} (e.g. UMLS tasks evaluated on COMETA which is SNOMED annotated), $7\%$ to $30\%$ higher R@1 is observed. \emph{Comb} task has the most balanced performance gains across all datasets.

\subsection{Analysis and Discussion }
\textbf{Task comparison.} To further understand the difference between synonym-based tasks and graph-based tasks, we illustrate qualitative examples in Table \ref{tab:case_review}. 
A model trained using the synonym task makes better predictions for scenarios involving medical synonyms and acronym (lines $1$, $2$). 
A model trained using the graph task performs better when mention and concept have an \texttt{is\_a} relationship (lines $3$, $4$).

\begin{figure}[t]
    \centering
    \includegraphics[trim=8pt 10pt 10pt 0pt, clip=true, width=0.32\textwidth]{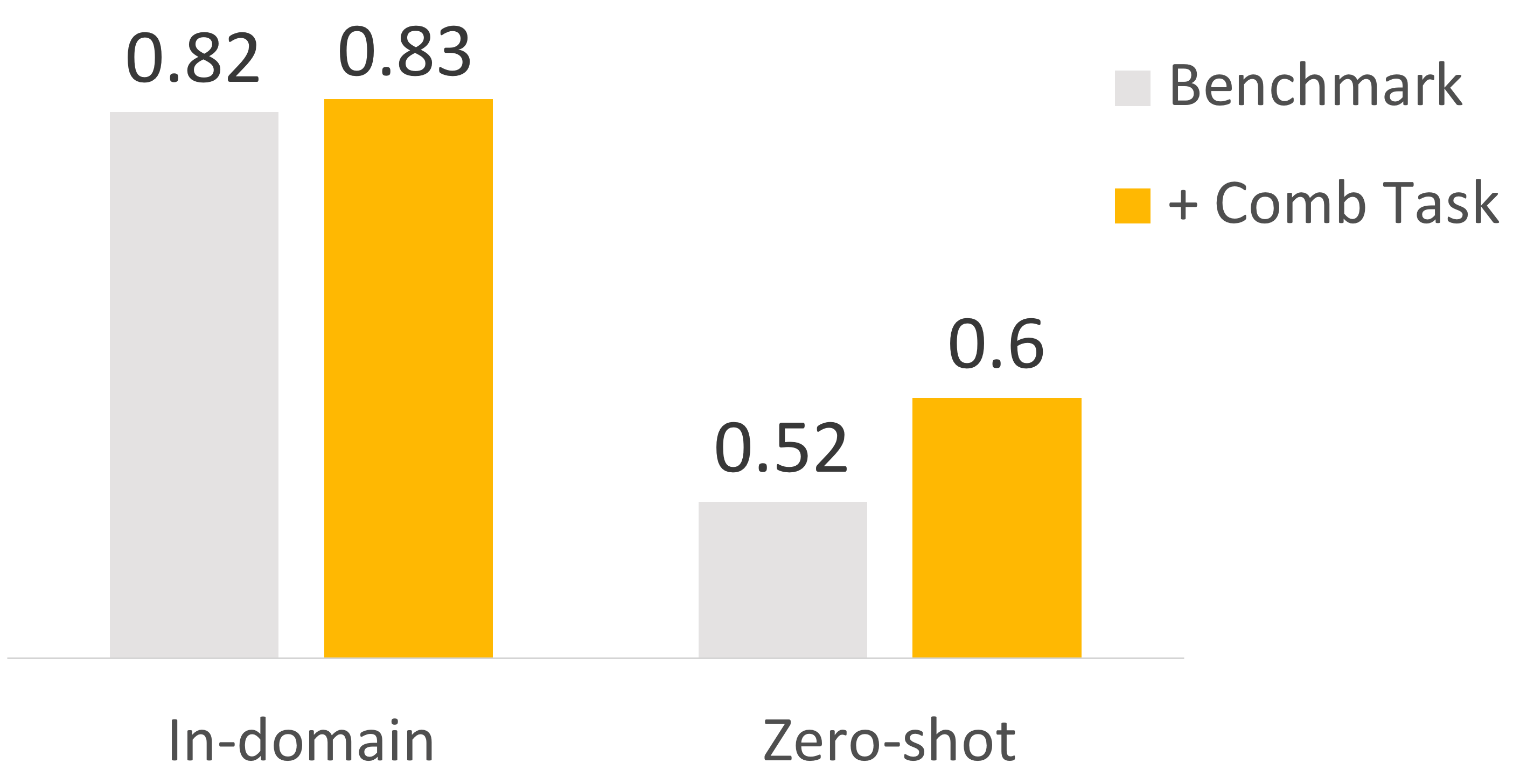}
    \caption{R@25. Using the \emph{comb} task as an auxiliary to the primary supervised loss, the model gains $1\%$ in-domain (3DNotes-SN) improvement , $8\%$ zero-shot (COMETA-ZS) improvement.}
    \label{fig:MTL}
\end{figure}

\textbf{Auxiliary task.} When annotations are available, our learning tasks can be used as an auxiliary to the primary loss. 
Using the 3DNotes-SN's annotated training set to train the primary supervised task, we set the \emph{comb} task as its auxiliary counterpart by summing the losses. 
We evaluate zero-shot performance on COMETA-ZS. We observe an $8\%$ increase in R@25, illustrated in Fig. \ref{fig:MTL}.
Since most annotations cover no more than a couple thousands concepts, which is a tiny portion of a typical medical KG's size, this demonstrates the generalizing capacity of our approach on the vast majority of unseen concepts. 

\textbf{Private KG.} In practice, if the target medical ontology is a private KG \cite{wise2020covid19, bhatia2020aws}, one can also consider customizing the learning tasks that follow the synonym and graph-based frameworks outlined in this work to bring greater gains. 

\section{Conclusion}
We present a framework for allowing entity retrieval models to mine rich semantics from a medical KG. 
We show its effectiveness in zero-shot settings through extensive experiments. 
In addition, we demonstrate the ease with which the framework can be adapted to serve as an auxiliary task when annotations are available. 
Future research should explore more fine-grained approaches to combine tasks.


\bibliographystyle{acl_natbib}
\bibliography{anthology,acl2021}


\end{document}